\documentclass[twoside,british,authoryear,preprint]{elsarticle}
\usepackage{lmodern}

\usepackage[T1]{fontenc}
\usepackage[latin9]{inputenc}
\usepackage{textcomp}
\usepackage{amsmath}
\usepackage{graphicx}

\makeatletter

\newcommand*\LyXZeroWidthSpace{\hspace{0pt}}
\newcommand{\noun}[1]{\textsc{#1}}
\providecommand{\tabularnewline}{\\}

\journal{Mechatronics}
\usepackage{textcomp}
\newcommand{\tc}{\ensuremath{\,\textrm{\textcelsius}}}
\newcommand{\ls}{\ensuremath{\,\ell\mathrm{s}^{-1}}}

\newcommand{\rlreward}{-1.2}

\newcommand{\bangbangavgcomfort}{55}

\newcommand{\rlcomfort}{67}

\newcommand{\bangbangavgenergy}{0.88}

\newcommand{\rlenergy}{0.77}

\newcommand{\bangbangavgrewardpctinc}{23}

\newcommand{\proportionalavgrewardpctinc}{43}

\newcommand{\commercialairrewardpctinc}{56}

\newcommand{\fuzzyetrewardpctinc}{40}

\newcommand{\bangbangavgenergypctdec}{13}

\newcommand{\fuzzyetenergypctdec}{37}

\newcommand{\bangbangavgcomfortpctinc}{23}

\newcommand{\rlepsilonzeropoint}{190\,000}
\newcommand{\rlgamma}{0.99}

\newcommand{\rllambda}{0.98}

\newcommand{\rlmaxstepspertrial}{500}
\newcommand{\rlmaxtrials}{200\,000}

\newcommand{\rlsimyears}{6.3}

\newcommand{\rltestepisodes}{200}

\newcommand{\rlbestalpha}{0.01}
\newcommand{\rlbestepsilon}{0.16}
\newcommand{\rlbestlambda}{0.98}

\newcommand{\venttemprange}{\left[7,60\right]}

\newcommand{\energyweight}{30\,000}

\newcommand{\cabintemprange}{\left[0,50\right]}
\newcommand{\blocktemprange}{\left[0,50\right]}
\newcommand{\ambienttemprange}{\left[0,40\right]}
\newcommand{\targetrange}{24\pm1\tc}
\newcommand{\maxctbtdiff}{30\tc}

\newcommand{\cooldowncabin}{45\tc}
\newcommand{\cooldownblock}{45\tc}
\newcommand{\cooldownambient}{20\tc}

\usepackage{color}

\makeatother

\usepackage{babel}
\begin{document}

\begin{frontmatter}{}

\title{Reinforcement Learning-based Thermal Comfort Control for Vehicle
Cabins}

\author[rvt]{J.~Brusey\fnref{fn1}\corref{cor1}}

\ead{j.brusey@coventry.ac.uk}

\author[rvt]{D.~Hintea\fnref{fn1}}

\author[rvt]{E.~Gaura\fnref{fn1}}

\author[rvt,focal]{N.~Beloe\fnref{fn2}}

\cortext[cor1]{Corresponding author}

\address[rvt]{Faculty of Engineering, Environment and Computing, Coventry University,
Gulson Rd, Coventry, West Midlands CV1 2JH, United Kingdom}

\address[focal]{Jaguar Land Rover Limited, Abbey Road, Whitley, Coventry, CV3 4LF,
United Kingdom}
\begin{abstract}
Vehicle climate control systems aim to keep passengers thermally
comfortable. However, current systems control temperature rather than
thermal comfort and tend to be energy hungry, which is of particular
concern when considering electric vehicles. This paper poses energy-efficient
vehicle comfort control as a Markov Decision Process, which is then
solved numerically using Sarsa($\lambda$) and an empirically validated,
single-zone, 1D thermal model of the cabin. The resulting controller
was tested in simulation using $\rltestepisodes$ randomly selected
scenarios and found to exceed the performance of bang-bang, proportional,
simple fuzzy logic, and commercial controllers with $\bangbangavgrewardpctinc$\%,
$\proportionalavgrewardpctinc$\%, $\fuzzyetrewardpctinc$\%, $\commercialairrewardpctinc$\%
increase, respectively. Compared to the next best performing controller,
energy consumption is reduced by $\bangbangavgenergypctdec$\% while
the proportion of time spent thermally comfortable is increased by
$\bangbangavgcomfortpctinc$\%.  These results indicate that this
is a viable approach that promises to translate into substantial comfort
and energy improvements in the car.\end{abstract}

\end{frontmatter}{ }

\section{Introduction}

Vehicle HVAC (Heating, ventilation, and air conditioning) systems
aim to ensure that passengers are thermally comfortable. Traditionally,
controllers for these systems are hand-coded and tuned to try to achieve
this goal. However, there are a number of drivers for change:
\begin{enumerate}
\item Current systems only control cabin temperature whereas thermal comfort
is also dependent on a multitude of other factors (such as radiant
heat and airflow). 
\item Past systems have relied on waste heat from the engine whereas electric
vehicles produce much less heat and so a different design is required.
\item Current systems are energy hungry whereas electric and hybrid vehicles
demand a much more energy efficient approach. \citet{Farrington2000}
report that air conditioning systems reduce the fuel economy of fuel-efficient
cars by about 50\%. 
\end{enumerate}
These drivers for change make redesign of many parts of the vehicle
comfort delivery system timely. As this comfort system design changes,
the controller must also adapt to best make use of the available actuation
options. 

The main idea in this paper is to show that Reinforcement Learning
(RL) reliably produces a controller that uses less energy while delivering
better comfort than existing hand-coded approaches (Section~\ref{sec:Evaluation}).
We also show that the trade-off between energy and comfort can be
adjusted to suit situations that demand either more comfort or better
energy efficiency (Section~\ref{sub:Effect-of-parameter}). The approach
requires a model of the cabin environment and we provide a simple,
empirically validated, lumped model of the cabin's thermal environment
(Section~\ref{sec:Thermal-Environmental-Model}). The problem is
then defined in terms of the state space (Section~\ref{sub:State-Representation}),
action space (Section~\ref{sub:Action-Representation}) and reward
function (Section~\ref{sub:Reward-Function}). Issues and implementation
ramifications of this approach are discussed in Section~\ref{sec:Discussion}.

\section{Related work}

\subsection{HVAC control methods in vehicles\label{sec:-Control-Methods}}

Much of the work on HVAC control~\citep{Stephen2010,Farzaneh2008,Atthajariyakul,Goenka2013}
remains rooted in thermal comfort models developed for home and office
indoor environments. The best known comfort model is the Predictive
Mean Vote (PMV) \citep{Fanger1973,Fanger1970,Gagge1967}, which estimates
comfort based on: environmental parameters (such as air temperature,
mean radiant temperature, relative air velocity and relative humidity);
and personal parameters (such as metabolic rate and clothing thermal
resistance). For example, \citet{Stephen2010} derive a PMV-based
fuzzy logic control mechanism, with rules like ``if temperature is
medium and activity is low, then PMV is near neutral''.

Although many aspects of vehicle thermal environment control are derivative
of that in buildings, the vehicle's thermal environment is transient
and non-uniform~\citep{Zhang2003}. Thus it is recognised that what
is appropriate in the thermal comfort model for a building may not
be appropriate in a car~\citep{Kranz2011,Croitoru2015}. 

While there are a number of thermal comfort models available, there
is disagreement between these models about what contribution different
parameters should have, or even what parameters to include~\citep{Croitoru2015}.
Moreover, there are clearly parameters that might be considered but
are not generally included. For example, occupants may enter the vehicle
with latent or stored heat, they may have a physiological condition
(such as a fever), or they may have cultural or personal preferences~\citep{Cheung2010}.
While there are many factors that can affect comfort, not all affect
it equally. While air temperature remains central to comfort, as the
number of sensors and intelligence of the controller within the car
increases, it becomes possible to include more factors.

A number of additional \emph{models}, \emph{estimators}, and \emph{predictors}
populate the literature, typically accompanied by a strategy for HVAC
control (e.g., \citet{Ueda1999} predicts comfort based on facial
skin temperature and cabin air temperature; \citet{Goenka2013} proposed
a zonal HVAC system driven on an occupant thermal comfort level based
on sensor measurements, thermal comfort charts, the ASHRAE thermal
scale, ISO 7730, the PMV index, the PPD index and their combination;
\citet{Kranz2011} applies artificial intelligence methods to extract
thermal comfort knowledge from the interaction between the passengers
and the HVAC controls). Not surprisingly, most, if not all, of the
proposed controllers are based on machine learning techniques. A prime
reason is that car cabin comfort control is non-linear with respect
to the observable state, for example: (a) the transfer of heat as
a function of vent speed and vent temperature is non-linear; (b) any
plant output limitation affects response in a non-linear fashion~\citep{Davis1998};
(c) comfort models, such as Predicted Mean Vote (PMV) and equivalent
temperature (ET), are a non-linear function of their inputs.

Fuzzy logic is a common HVAC control approach given the imprecise
nature of comfort~\citep{Davis1996,Davis1998,Beinarts2011,Singh2006,Thompson,Stephen2010,Nasution2008,Farzaneh2008,Gach1997}
and many fuzzy-logic controllers have been found to perform better
than the traditional air temperature controllers. \citet{Farzaneh2008}
demonstrated that even better results were obtained when the parameters
of the comfort oriented fuzzy controller were optimised by a genetic
algorithm. Such controllers are, however, computationally expensive
and can be difficult to design.

\subsection{Reinforcement learning-based control applications\label{sec:Reinforcement-Learning-Theory}}

\citet{Dalamagkidis2007} and \citet{Fazenda2014} have examined the
problem of optimising HVAC thermal comfort-based control through a
RL-based technique in the context of buildings rather than cars. \citet{Dalamagkidis2007}
developed and simulated a reinforcement learning-based controller
using Matlab/Simulink. The reward is a function of the building occupants'
thermal comfort, the energy consumption and the indoor air quality.
The proposed controller was compared to a Fuzzy-PD controller and
a traditional on/off controller (an evaluation approach also applied
here). The results showed that, after a couple of simulated years
of training, the reinforcement learning-based controller performed
better in comparison to the other two controllers.

\citet{Dalamagkidis2007} highlight an issue with regard to reinforcement
learning-based controllers---that of sufficient exploration. Taking
random actions, even during short times, is unacceptable for a system
deployed in a real environment and the authors recommend to exhaustively
train the controller prior deployment and allow minimal or no exploration
at all afterwards. This work provided inspiration and a good foundation
for our work in vehicle cabins.

\citet{Fazenda2014} have examined the problem of optimising comfort
and energy using Q-learning with a state space that includes the time
of day. They break the control problem down into: bang-bang control
(when to turn the heater on or off) and set-point control (what temperature
to request at what time). In their work, the tenant immediately responds
to discomfort, which might seem unrealistic, but it provides similar
input to the thermal comfort model used here. By including time, they
neatly provide for pre-heating or cooling and this approach might
also be used for the car cabin. 

Less recently, \citet{Anderson1997} have examined the problem of
a simulated heating coil and combined a PI (proportional-integral)
controller with an RL supervisor. They showed that the combined approach
outperforms the base PI controller. This combination is similar to
the approach here where the RL action is a vent temperature set-point
that is passed to a base controller to achieve. 

\citet{Dounis2009} provide a detailed review of computational intelligence
approaches in the built environment and show that, for the built environment,
a variety of adaptive control approaches have been tried and advanced
approaches (such as RL) have led to improved comfort and energy savings. 

This past work demonstrates that RL, while untested, may be appropriate
in this domain.

\section{Materials and methods\label{sec:materials-and-methods}}

\begin{figure}
\includegraphics{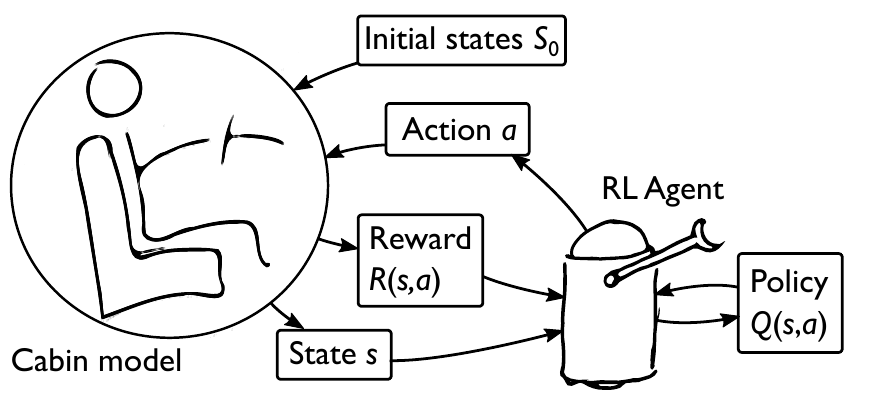}

\noindent \centering{}\protect\caption{The process of finding an optimal policy with RL involves modelling
the cabin environment $T$, identifying the state $S$ and action
$A$ spaces, defining the distribution $S_{0}$ of initial states,
and defining an appropriate reward function $\mathcal{R}(s,a)$.\label{fig:Overall-system-architecture-sim} }
\end{figure}

We formulate the cabin comfort control problem as a Markov Decision
Process (MDP) with continuous states defined by the tuple $\langle S,S_{0},A,T,\mathcal{R},\gamma\rangle$,
where $S$ is the (infinite) set of states of the cabin environment
from which a set of initial states $S_{0}\subseteq S$ is drawn, $A$
is a finite set of actions (e.g., setting the blend door position),
$T\colon S\times A\to S$ is a deterministic environmental model that
maps states and actions to subsequent states, $\mathcal{R}\colon S\times A\to\Re$
is a function expressing the reward for taking an action in a particular
state, and $\gamma$ is a discount factor such that, for $\gamma<1$,
a reward achieved in the future is worth less than a reward achieved
immediately. 

The solution of the MDP is a policy $\pi\colon S\to A$ or mapping
from states to actions and, in particular, an optimal solution is
one that maximises the long-term, discounted expected reward. In algorithms
such as Q-learning and Sarsa($\lambda$), rather than find the policy
directly, we estimate the expected value or utility $Q^{\pi}\left(s,a\right)$
of each state, action combination when following policy $\pi$. This
expected value is the immediate reward $\mathcal{R}\left(s,a\right)$
plus the discounted subsequent reward, which can thus can be defined
recursively,
\begin{equation}
Q^{\pi}\left(s,a\right)=\mathcal{R}\left(s,a\right)+\gamma Q^{\pi}\left(T\left(s,a\right),\pi\left(T\left(s,a\right)\right)\right).\label{eq:qsa}
\end{equation}
We can then progress greedily towards the optimum policy by updating
the policy $\pi$ to be that which maximises $Q^{\pi}$, or, 
\begin{equation}
\pi\left(s\right)\leftarrow\arg\max_{a\in A}Q^{\pi}\left(s,a\right).\label{eq:policy}
\end{equation}
Since the policy for any state is easy to calculate from $Q^{\pi}$,
it does not need to be explicitly stored.

For finite state MDPs, algorithms such as Monte Carlo Exploring Starts
(MCES) and Monte Carlo $\varepsilon$-soft~\citet[§5.3,5.4]{Sutton1998}
use repeated application of (\ref{eq:qsa}) and (\ref{eq:policy})
to converge on the optimal policy. To avoid getting stuck in a local
minima, they include some random exploration and this is sufficient
to ensure that they always converge on the global optimum policy.
For continuous state MDPs, $Q^{\pi}\left(s,a\right)$ must be approximated
using a function $f\left(\vec{\theta},s,a\right)$ parameterised by
a vector $\vec{\theta}$ and algorithms, such as Sarsa$\left(\lambda\right)$,
that use this approach may not converge on the optimum policy but
may oscillate~\citep{Gordon2000}. 

A learning \emph{episode} begins by selecting an initial state at
random from the distribution $s_{0}\sim S_{0}$ and then continues
with the agent selecting an action and the cabin model returning a
new state and reward until a maximum number of steps is reached. For
some problems, it is possible to have a terminal state that ends the
episode. However, this is not possible here, since the reaching comfort
is not sufficient; the agent needs to efficiently maintain comfort
as well. The initial state distribution should be comprehensive to
avoid leaving parts of the state space unexplored. The agent is $\varepsilon$-greedy,
which means that with probability $\varepsilon$ it selects a random
action and otherwise it selects according to the largest estimated
utility for that state, as per (\ref{eq:policy}). 

Although it might be possible to implement a learning system directly
in the car, prior works in this domain (such as, \citet{Ng2006})
suggest learning in simulation first. In principle, the learnt policy
can then be implemented in the car cabin either as a fixed policy
or as a start point for continued learning. In this work, we only
examine the system in simulation and implementing in the car is left
to future work. 

Given this basis for learning, we now define each aspect of the MDP,
beginning with the model.

\subsection{Cabin Thermal Environmental Model\label{sec:Thermal-Environmental-Model}}

Car cabin thermal modelling has been investigated by a number of authors~\citep{Lee2015,Torregrosa-Jaime2015},
typically to examine the trade-off between comfort and energy use.
Simple 1D models are appropriate for optimisation (e.g., \citet{Lee2015}
examines the effect of different coolant fluids) since they allow
the consequences of changes to be quickly evaluated. Some simplifying
assumptions are necessary and different works tend to make different
assumptions about the cabin environment. For example, \citet{Lee2015}
include the effect of engine heat on supply and return ducts, whereas
\citet{Torregrosa-Jaime2015} include radiant heat effects for a multi-zone
minivan. Our focus here is to provide a clearly described, simple
model that might be expanded upon but which is validated against data
from a real car in a climatic wind tunnel (Section~\ref{sub:Model-validation}). 

\begin{figure}
\includegraphics{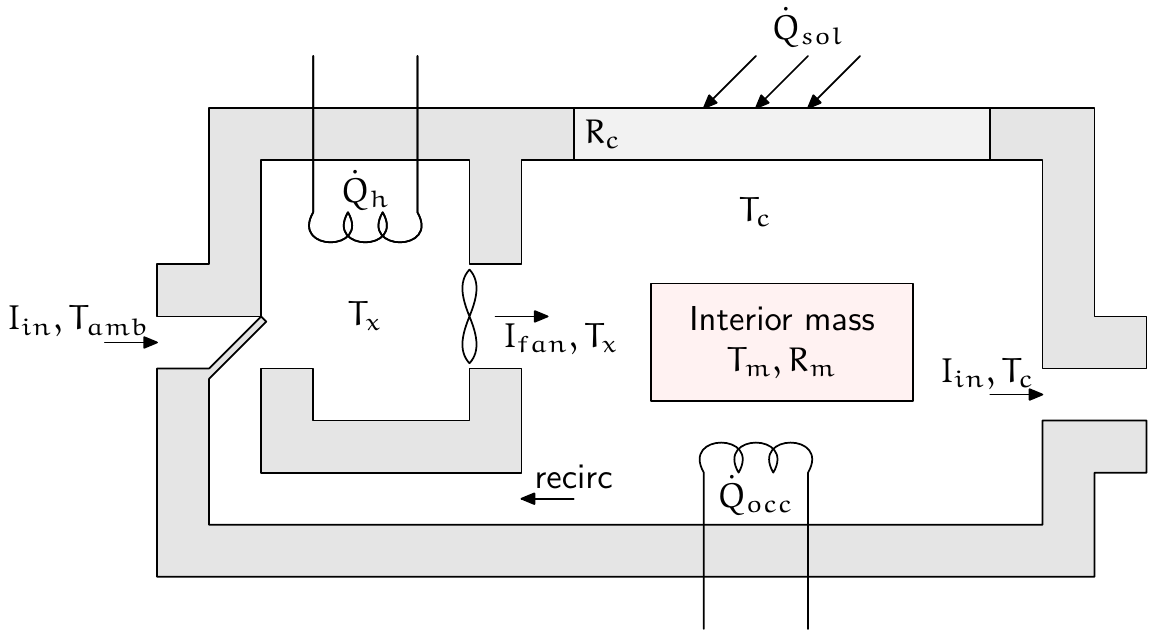}

\protect\caption{Schematic of the simplified cabin model used for learning a controller.\label{fig:Simplified-cabin-model}}
\end{figure}

Our simple cabin model is shown in Figure~\ref{fig:Simplified-cabin-model}
and this corresponds to a system of three heat balance equations (heat
in $=$ heat out $+$ heat stored), 
\begin{eqnarray}
\dot{Q}_{h}+I_{\mathit{in}}\left(T_{\mathit{amb}}-T_{c}\right) & = & I_{\mathit{fan}}\left(T_{x}-T_{c}\right)\label{eq:hb_x}\\
I_{fan}\left(T_{x}-T_{c}\right)+\dot{Q}_{sol}+\dot{Q}_{occ}+\frac{T_{m}-T_{c}}{R_{m}} & = & \frac{T_{c}-T_{\mathit{amb}}}{R_{c}}+C_{c}k\frac{dT_{c}}{dt}\label{eq:hb_c}\\
\frac{T_{c}-T_{m}}{R_{m}} & = & C_{m}\frac{dT_{m}}{dt}\label{eq:hb_m}
\end{eqnarray}
where $\dot{Q}$ is the change in heat energy, $I$ is the current
(or mass flow of heated air), $T$ is the temperature, $R$ is the
thermal resistivity, and $C$ is the thermal capacitance. Subscripts
are: $h$ heat pump, $\mathit{in}$ input air, $\mathit{amb}$ ambient
air, $c$ cabin air, $\mathit{fan}$ blower fan, $x$ mixed air, $\mathit{sol}$
solar load, $\mathit{occ}$ occupant, and $m$ interior mass. A cabin
capacitance factor $k$ is used to account for the difference between
the experimentally observed capacitance of the cabin air and the theoretical
thermal capacitance of air. This difference is probably due factors
such as the air mixing time (which is otherwise assumed to be instantaneous
in the model). The recirculation factor $\alpha=I_{in}/I_{\mathit{fan}}$
corresponds to the percentage of fresh air. Note that the mixing chamber
heat storage is assumed to be negligible. For the purposes of this
work, we take the work done (or energy consumed) by the HVAC to be
simply $W_{h}=\left|\dot{Q}_{h}\right|$ and ignore the energy cost
for the fan.

Model constants, shown in Table~\ref{tab:Model-constants}, were
selected to best match the target car, a Jaguar model XJ sedan. This
car was used for model validation in Section~\ref{sub:Model-validation}.

\begin{table}
\protect\caption{Model constants\label{tab:Model-constants}}

\begin{tabular}{lr}
\hline 
Cabin volume $V_{c}$ & $2.5\,\mathrm{m^{3}}$\tabularnewline
Cabin capacitance factor $k$ & $8$\tabularnewline
Solar load $\dot{Q}_{sol}$ & $150\,\mathrm{W}$\tabularnewline
Occupant load $\dot{Q}_{occ}$ & $120\,\mathrm{W}$\tabularnewline
Cabin resistivity $R_{c}$ & $1/\left(5.741626794\times4.0\right)\,\mathrm{K}.\mathrm{W}^{-1}$\tabularnewline
Interior mass resistivity $R_{m}$ & $1/\left(75\times1.08\right)\,\mathrm{K}.\mathrm{W}^{-1}$\tabularnewline
Interior mass capacitance $C_{m}$ & $450\times0.02\times7850\,\mathrm{J}.\mathrm{K}^{-1}$\tabularnewline
\hline 
\end{tabular}
\end{table}

It is assumed that there is no air leakage. Nor is the vehicle velocity
taken into account. In comparison with \citet{Lee2015} this model
does not deal with the internals of the evaporator but rather considers
the combined heat sum from a heat pump. Also, heat effects from the
internal combustion engine (through the firewall or supply ducts)
are not considered here (and may be inappropriate for an electric
vehicle). \citet{Torregrosa-Jaime2015} have a more sophisticated
model that includes two zones for a minivan. In comparison to the
work here, they include radiative heat transfer between cabin walls
and the interior mass as well as between the cabin walls and the sky.

\subsection{Model validation\label{sub:Model-validation}}

The simulation data was compared to empirical data collected by the
authors within various warm-up and cool-down scenarios (described
in \citet{Hintea2014}). 
\begin{figure}
\includegraphics[scale=0.75]{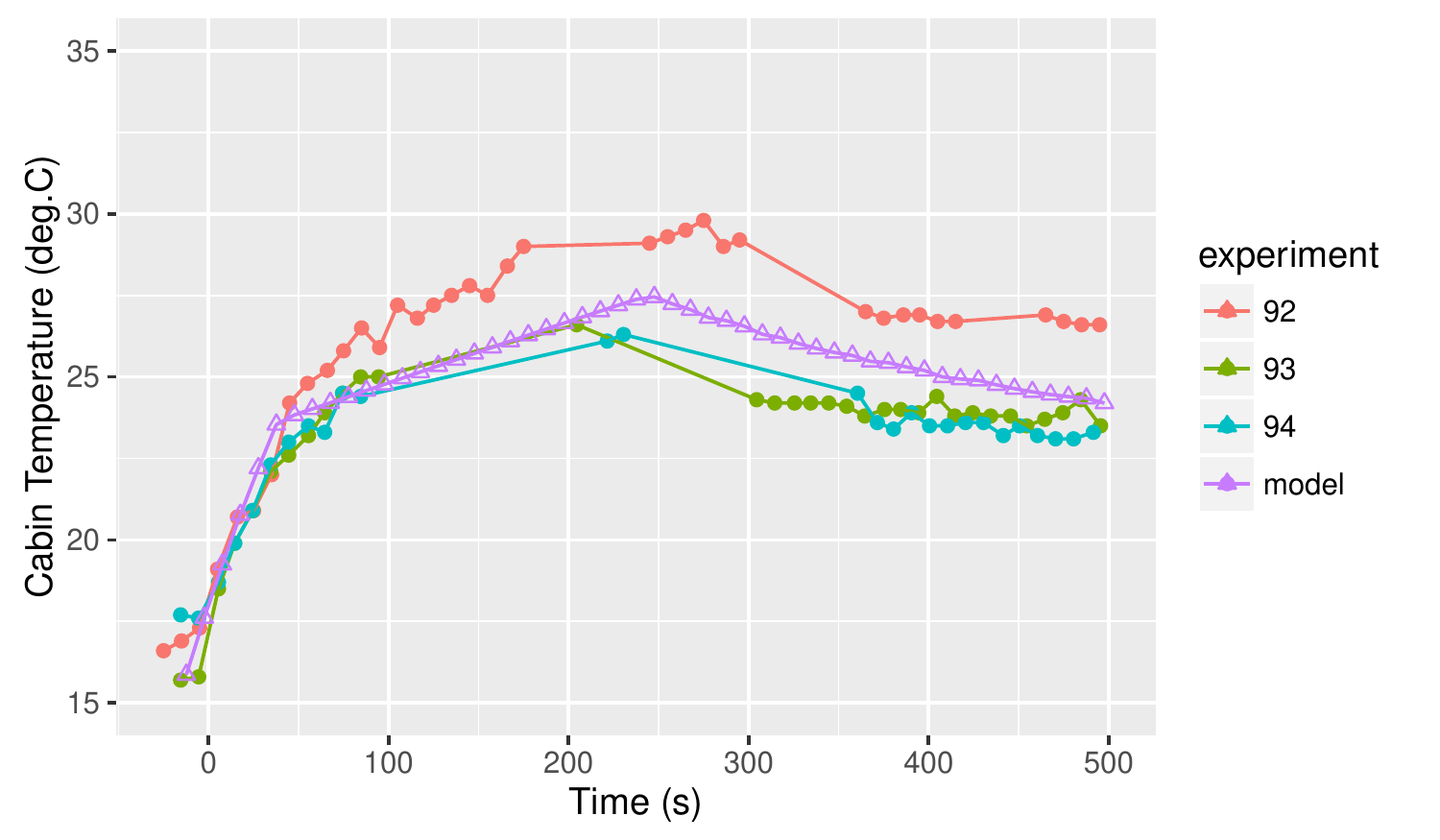}

\protect\caption{Car cabin warm-up experiment showing real and simulated (denoted `model')
results time-aligned at $18\tc$.\label{fig:Car-cabin-warm-up} The
time-series shows overshoot in the controller, probably due to lag
in the in-car sensor. The proportional controller with an averaged
sensor (see Section~\ref{sub:Bang-bang-controller}) is used with
the simulated model and this produces a similar overshoot. }
\end{figure}
\begin{figure}
\includegraphics[scale=0.75]{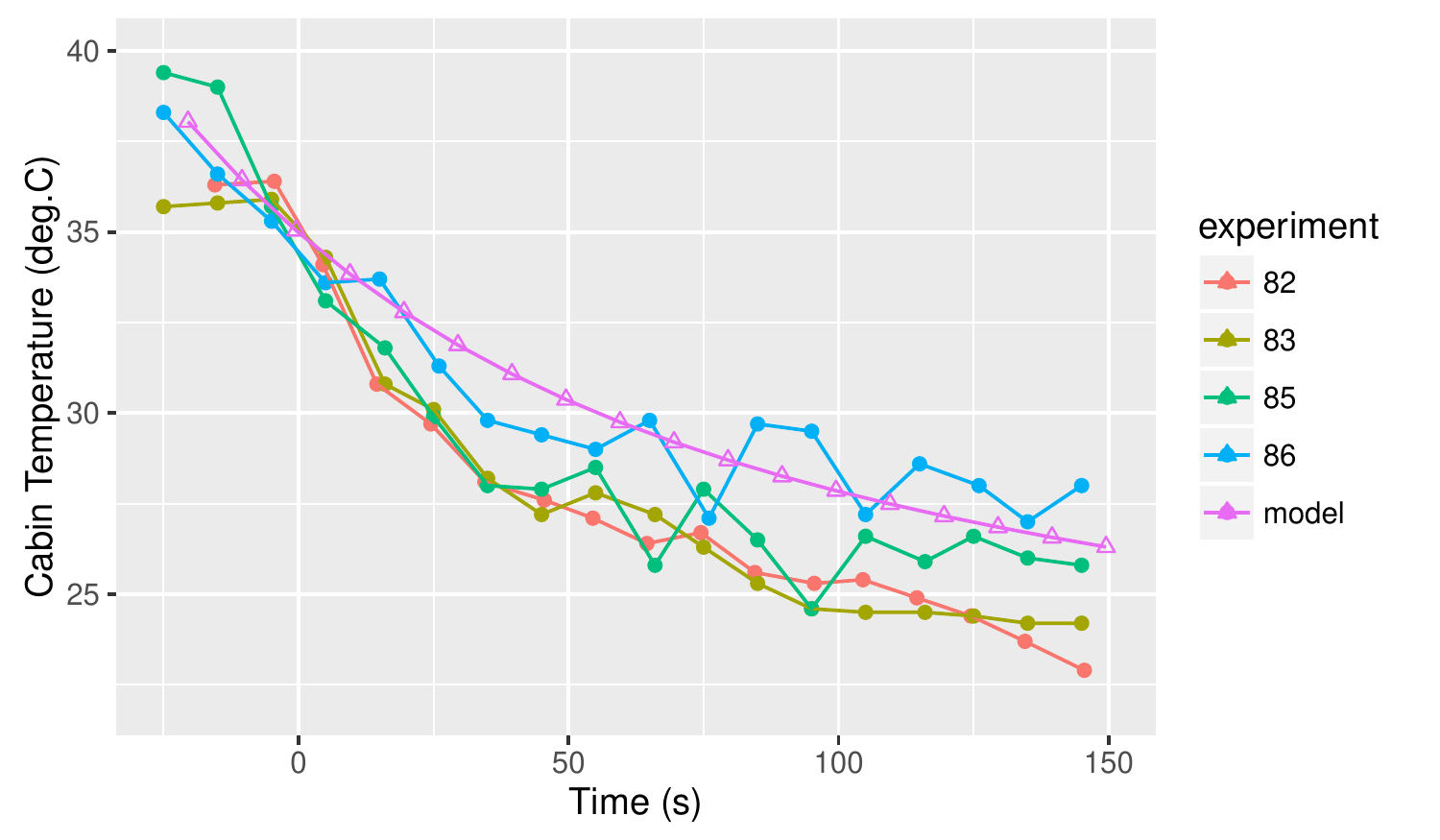}

\protect\caption{Car cabin cool-down experiment showing real and simulated (denoted
`model') results time-aligned at $35\tc$.\label{fig:Car-cabin-cool-down}}
\end{figure}
Figures~\ref{fig:Car-cabin-warm-up} and~\ref{fig:Car-cabin-cool-down},
based on experiments with a Jaguar XJ in MIRA LTD's climatic wind
tunnel, show warming (from cold) and cooling (from hot) the car cabin
based on head-rest height temperature sensors over a number of experiments
and also showing the simulated or `model' results. Simulated results
are based on the bang-bang controller described further in Section~\ref{sub:Bang-bang-controller}.
These graphs demonstrate that the simulation broadly matches the characteristics
of the physical system and thus that a controller that performs well
with the simulation is likely to work well in practise in terms of
control of temperature. 

Although the modelling of energy use is based on reasonable assumptions,
at this stage we have no experimental data with which to validate
the model. Energy use is difficult to estimate precisely in practise
since, for example, latent heat from the engine is used to heat the
cabin. Thus energy use in practise may differ from the simulation.

\subsection{State representation\label{sub:State-Representation}}

The state of the cabin environment is a vector comprising: the cabin
air temperature $T_{c}$, the interior mass temperature $T_{m}$ and
the outside air temperature $T_{\mathit{amb}}$. Equivalent temperature
(ET) is not an explicit component of the state but is computed using
a formula (referring to sedentary conditions only, that is energy
metabolism~$<70\,\mbox{Wm}^{-2}$) introduced by \citet{Madsen1978,Madsen1984},
\begin{equation}
T_{e}=\begin{cases}
0.5(T_{c}+T_{r}), & \mbox{for air flow}\,\dot{v}_{c}\leq0.1\,\mbox{ms}^{-1}\\
0.55T_{c}+0.45T_{r}+\frac{0.24-0.75\sqrt{\dot{v}_{c}}}{1+I_{cl}}\left(36.5-T_{c}\right), & \mbox{for}\,\dot{v}_{c}>0.1\,\mbox{ms}^{-1}
\end{cases}
\end{equation}
The air flow corresponding to the cabin occupant $\dot{v}_{c}$ is
not directly available and it is estimated here by dividing the vent
air flow $v_{i}$ by $10$. The value was selected based on cabin
air flow measurements in the literature~\citet{Neacsu2011}. The
mean radiant temperature $T_{r}$ is assumed to be equal to the interior
mass temperature $T_{m}$. For this work, the clothing insulation
$I_{cl}$ is set to a constant value of $0.7$~clo corresponding
to \LyXZeroWidthSpace long trousers and short sleeve, light-coloured
blouse or shirt. Note that ET is provided as input to the `et' variants
of the hand-coded controllers but is not explicitly provided to the
RL controller. 

At the beginning of each controller training episode, the initial
state vector is selected at random from a uniform distribution over
the full range of values for each of:
\begin{itemize}
\item \noindent Interior temperature $T_{m}$: $\blocktemprange\tc$.
\item \noindent Outside temperature $T_{o}$: $\ambienttemprange\tc$.
\item \noindent Cabin air temperature $T_{a}$: $\cabintemprange\tc$. 
\end{itemize}
The representation of the state is minimal, sufficient (along with
the action) for the reward function, and Markovian (in terms of the
simulation). Some elements that are held constant in this model (such
as the solar load) might also be included in the state vector if they
were allowed to vary. 

Selection of the initial state and range of states is influenced by
the episode length and what is likely to occur. Episode length places
a limit on the extreme values. For example, it might take more than
$\rlmaxstepspertrial$ steps to achieve a comfortable state from a
very high or low start temperature. From such a start point, any policy
looks equally bad. 

We also eliminate start states where the interior mass temperature
is different from the cabin temperature by more than $\maxctbtdiff$,
as this situation is considered to be unlikely.

Function approximation is used by the Sarsa($\lambda$) algorithm
to avoid having to discretise the state and also to support a large
state space. Function approximation involves defining a parameter
vector $\mathbf{\theta}=\left(\theta_{1},\theta_{2},\ldots\right)^{T}$
thus allowing $Q$ to be approximated by a smooth function
\[
\hat{Q}\left(s,a\right)=f_{\theta}\left(s,a\right).
\]
The function approximator used in this case is tile coding  and the
configuration of the function approximator is further below.

\subsection{Tile coding\label{sub:Policy}}

The tile coding parameters used for this problem are presented in
Table~\ref{tab:Tile-coding-parameters}. In contrast with other work,
rather than use a separate function approximator for each action,
a single function approximator is used with tiles that span the combined
state and action spaces. The tile coding used to represent the action-values
included 30 tiles, 10 tiles integrating variables $\left(T_{a},T_{m},T_{o},T_{i},v_{i},A_{r}\right)$
and 20 tiles integrating variables $\left(T_{a},T_{i},v_{i},A_{r}\right)$.
Note that ET is not included (since it is not part of the state vector).
\begin{table}
\protect\caption{Tile coding parameters used to learn the control policy.\label{tab:Tile-coding-parameters}}

\noindent \centering{}%
\begin{tabular}{cccc}
Variable & Minimum & Maximum & Intervals\tabularnewline
\hline 
$T_{a}$ & $0$ & $50$ & $26$\tabularnewline
$T_{m}$ & $10$ & $40$ & $7$\tabularnewline
$T_{o}$ & $0$ & $40$ & 7\tabularnewline
$T_{i}$ & $0$ & $60$ & 3\tabularnewline
$v_{i}$ & 1 & 100 & 3\tabularnewline
$A_{r}$ & 0 & 1 & 3\tabularnewline
\end{tabular}
\end{table}

\subsection{Action representation\label{sub:Action-Representation}}

The set of actions consists of a vector $\left(v_{i},T_{i},A_{r}\right)^{T}$
where each component of the vector takes on one of a small set of
discrete values. Specifically, there are four possible vent air flows
$v_{i}\in\left\{ 1,34,67,100\right\} \ls$. Five possible vent air
temperatures can be selected, which are evenly defined over the range
$T_{i}\in\venttemprange\tc$. Lastly, three recirculation flap positions
are available $A_{r}\in\left\{ 0,\frac{1}{2},1\right\} $. This yields
a total of 60 ($4\times5\times3$) possible actions.

\subsection{Reward function\label{sub:Reward-Function}}

The learning goal is to maximise the time spent in comfort (defined
here as when the occupant ET is $\targetrange$) while minimising
energy use. This can be expressed as the reward function,

\begin{eqnarray}
\mathcal{R}\left(s,a\right) & = & \mathcal{R}_{C}\left(s\right)-E\left(s,a\right)/w\\
\mathcal{R}_{C}(s) & = & \begin{cases}
0 & \mbox{if }T_{e}\in\targetrange\\
-1 & \mbox{otherwise}
\end{cases}\\
E(s,a) & = & \left|\dot{Q}_{E}\right|+2v_{i}
\end{eqnarray}
where $\mathcal{R}_{C}$ is the penalty for being uncomfortable, $E$
is the energy cost, $w=\energyweight$ is the energy weight divisor
(which, in lay terms, means that a 1\% improvement in comfort is equivalent
to $300\,\mathrm{W}$). This weight can be adjusted to give a different
trade-off between energy and comfort (see Section~\ref{sub:Effect-of-parameter}).
 The above reward function could be further extended to include goals
such as minimising fan noise or keeping the screen clear. Illegal
states (where component values are out of bounds) are not explicitly
penalised but act as an absorbing state with worst case penalty, which
is sufficient to ensure that the learning agent avoids them.

\subsection{Meta parameters\label{sub:Parameter-values}}

Meta parameters control the learning process and may affect how quickly
learning proceeds. The first is the number of steps per episode, which
is set at $\rlmaxstepspertrial$. This allows the agent to reach a
comfortable state from any start state but also that the episode length
is not so long that new start states are rarely experienced. The reward
discount factor $\gamma=\rlgamma$ ensures that a policy is appropriately
rewarded for actions that do not produce immediate reward. Given that
the reward function does not give reward for moving towards comfort
(but only for reaching it), setting $\gamma$ close to 1 allows the
agent to learn to achieve comfort even from extreme initial temperatures.
The learning rate $\alpha=\rlbestalpha$, exploration factor $\varepsilon=\rlbestepsilon$
(for first $\rlepsilonzeropoint$ episodes and zero thereafter), and
eligibility trace decay $\lambda=\rllambda$ were decided by looking
at the performance over the first $2\,000$ episodes, as discussed
in Section~\ref{sub:Effect-of-parameter}.

\subsection{Evaluation method}

The performance of the RL controller is tested using a set of $\rltestepisodes$
randomly pre-selected start states $S_{T}\subset S_{0}$ at regular
intervals during learning. This set is referred to as the \emph{test
scenario set}. This approach provides a standard test that can be
used for all controllers to provide fair comparison while ensuring
that the test is reasonably comprehensive over possible start states.

\section{Evaluation\label{sec:Evaluation}}

The RL-based controller is evaluated by comparing its performance
with: a bang-bang controller, a proportional controller, a commercial
controller, and a fuzzy-logic controller. For each controller, three
possible temperature sensors $T_{s}$ are simulated: the true cabin
air temperature (air), the average of cabin and interior mass temperatures
(avg), and the equivalent temperature (et). All controllers actuate
as per the action representation (see Section~\ref{sub:Action-Representation}).

\subsection{Bang-bang, proportional and commercial controllers\label{sub:Bang-bang-controller}}

The first three hand-coded controllers are somewhat similar. The bang-bang
controller blows the maximum fan rate to cool or warm the cabin until
it is within $1\tc$ of the target, at which point it blows the minimum
fan rate and tries to match the target temperature. The proportional
controller is similar but it reduces the fan speed exponentially $v_{i}=100-99\exp\left(-\frac{\left|T_{s}-24\right|}{10}\right)$
as the sensor temperature nears the target temperature. The commercial
controller is based on a commercial specification. This tends to use
lower fan rates than the proportional controller, probably to avoid
noise and vibration, but is otherwise quite similar.

\subsection{Simple fuzzy logic controller\label{sub:Fuzzy-Logic}}

For the evaluation here, a simple fuzzy logic controller was implemented
in Java using the fuzzylite library version~1.0~\citep{fl::fuzzylite}.
Apart from the sensor temperature $T_{s}$ , this controller also
receives interior mass temperature $T_{m}$. Fuzzy set membership
functions for input temperatures $T_{s},T_{m}$ are\noun{ neutral}
($\targetrange$), \noun{cold} (below \noun{neutral}) and \noun{hot
}(above) with some ramped overlap between each range. For vent temperature,
the sets are \noun{low} (below $10\tc$), \noun{medium} (around $20\tc$),
\noun{high} (above $30\tc$) and for vent flow rate, \noun{low }(below
$30\ls$), \noun{medium} (around $50\ls$), and \noun{high} (above
$70\ls$) with similar ramped overlaps.

\begin{figure}
\includegraphics{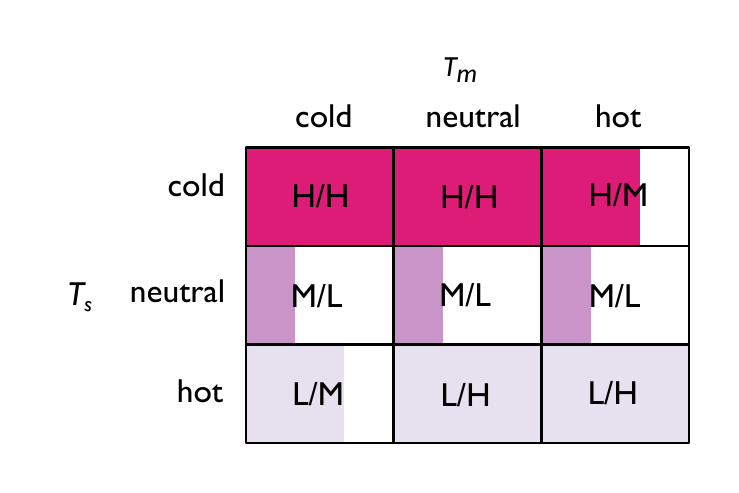}

\protect\caption{Simple fuzzy controller rules expressed as a table with outputs $T_{i}$/$v_{i}$.
The two inputs (sensor $T_{s}$ and interior mass temperature $T_{m}$)
are used to derive control of the vent temperature $T_{i}$ (\noun{high},
\noun{medium} or \noun{low}) and fan speed $v_{i}$ (\noun{high},
\noun{medium}, or \noun{low}). E.g., if $T_{s}$ is \noun{hot} and
$T_{m}$ is \noun{cold}, then $T_{i}$ is set \noun{low} and $v_{i}$
is set \noun{medium}.\label{fig:Fuzzy-rules-expressed}}
\end{figure}

The fuzzy logic rules are summarised in Figure~\ref{fig:Fuzzy-rules-expressed}.
These rules are slightly modified from those used by \citet{Dalamagkidis2007}
and \citet{Kelly2003}.

\subsection{Results\label{sec:Results}}

\begin{figure}
\noindent \includegraphics[scale=0.8]{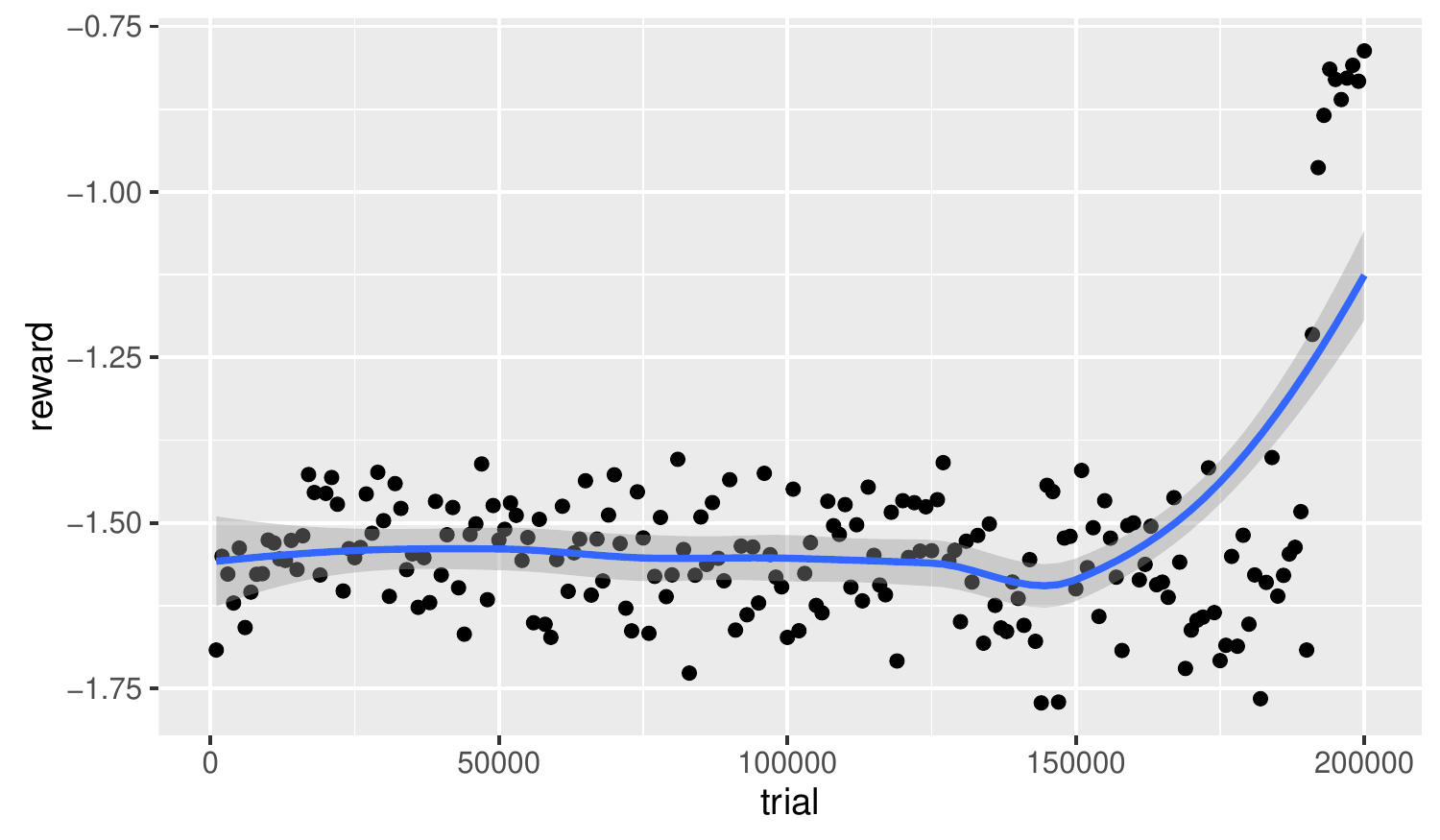}\protect\caption{Policy performance during learning for Sarsa($\lambda$). The first
$\rlepsilonzeropoint$ episodes are with exploration $\varepsilon=\rlbestepsilon$
while the rest are with no exploration $\varepsilon=0$. A LOESS fit
of the reward (with shaded 0.95 confidence interval) is also shown.
\label{fig:Policy-performance}}
\end{figure}

The relative performance of the RL controller compared with that of
the hand-coded controllers is given in Table~\ref{tab:Reward-Comfort-and}.
\begin{table}
\protect\caption{Reward, comfort and energy performance of the controllers over the
test scenario set for commercial, bang-bang, proportional, fuzzy logic
and RL agents. Sensors for the manual agents are cabin temperature
(air), an average of cabin and interior mass (avg), or equivalent
temperature (et).\label{tab:Reward-Comfort-and}}

\centering{}\begin{tabular}{p{0.25\columnwidth}p{0.2\columnwidth}p{0.2\columnwidth}p{0.2\columnwidth}}
Agent & Average reward & \% Time Spent in Comfort & Average HVAC power\tabularnewline
\hline
commercial-avg & $-2.9$ & $5$\% & $1.4$~kW\tabularnewline
commercial-et & $-2.9$ & $5.1$\% & $1.4$~kW\tabularnewline
bang-bang-et & $-2.8$ & $28$\% & $2$~kW\tabularnewline
commercial-air & $-2.8$ & $6.6$\% & $1.3$~kW\tabularnewline
fuzzy-avg & $-2.7$ & $2.6$\% & $0.94$~kW\tabularnewline
fuzzy-air & $-2.5$ & $2.2$\% & $0.76$~kW\tabularnewline
proportional-et & $-2.3$ & $18$\% & $0.91$~kW\tabularnewline
bang-bang-air & $-2.3$ & $13$\% & $0.72$~kW\tabularnewline
proportional-air & $-2.2$ & $12$\% & $0.59$~kW\tabularnewline
proportional-avg & $-2.2$ & $17$\% & $0.7$~kW\tabularnewline
fuzzy-et & $-2.1$ & $42$\% & $1.2$~kW\tabularnewline
bang-bang-avg & $-1.6$ & $55$\% & $0.88$~kW\tabularnewline
rl & $-1.2$ & $67$\% & $0.77$~kW\tabularnewline
\hline\end{tabular}
\end{table}
The RL controller gives the largest (least negative) average per-step
reward, uses less power and provides more comfort. This performance
evolves during learning as shown in Figure~\ref{fig:Policy-performance}.
The RL controller achieves an average reward of $\rlreward$ after
$\rlmaxtrials$ learning episodes (approximately $\rlsimyears$ simulated
years). Learning for the Sarsa($\lambda$) algorithm (implemented
in Java), corresponding to $\rlmaxtrials$ episodes, completed in
85 minutes on a 2.9~GHz Intel\textregistered{} Core\texttrademark{}
i7 processor. 

These results translate into an average factor of $\fuzzyetenergypctdec$\%
energy reduction over the test scenarios set when compared to the
simple fuzzy logic-based controller, while thermal comfort was achieved
and maintained successfully. 

\begin{figure}
\centering{}\includegraphics[scale=0.8]{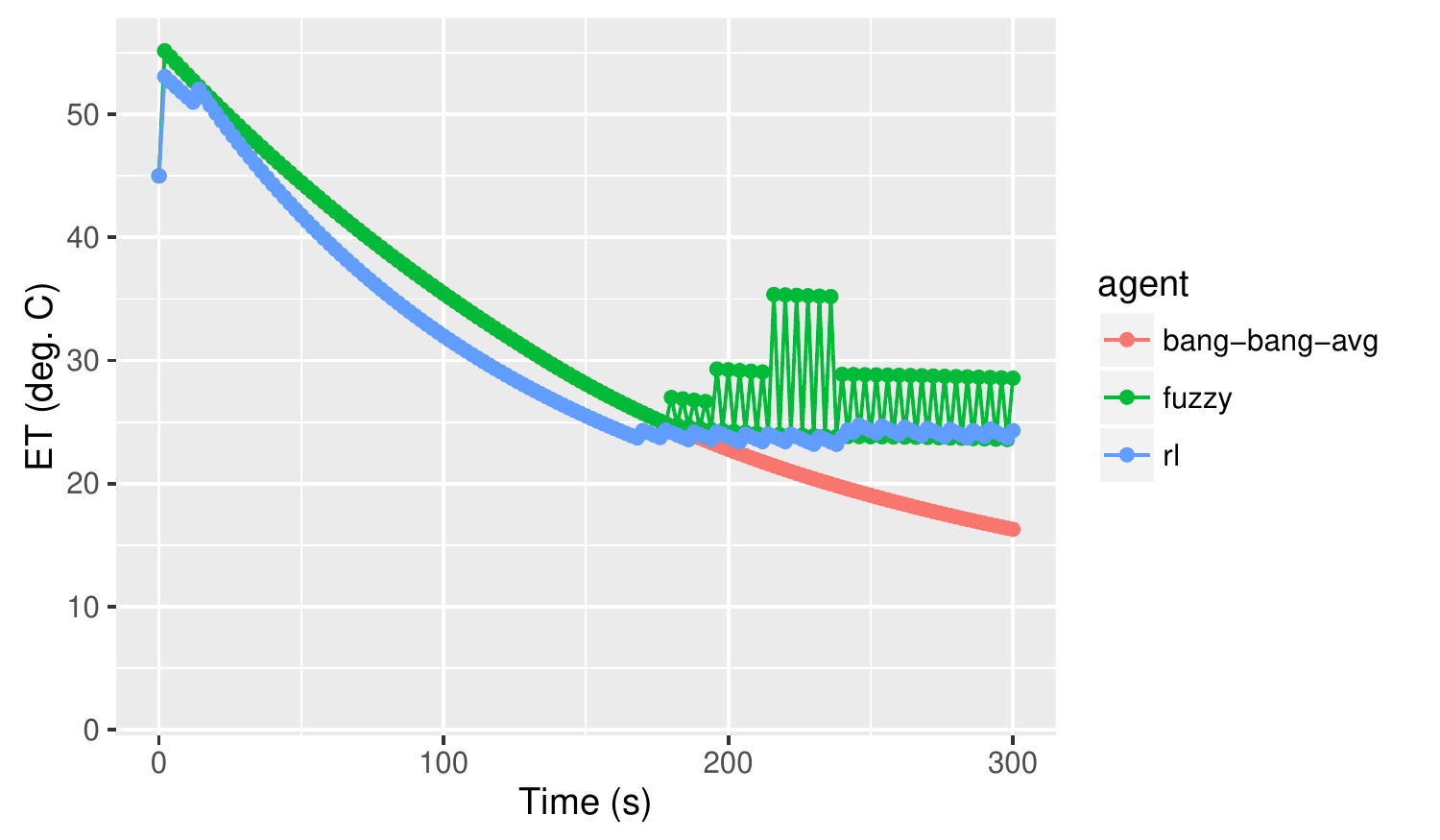}\protect\caption{Comparison of how each agent responds to an initially warm cabin.
\label{fig:Scenario-1}}
\end{figure}

Figure~\ref{fig:Scenario-1} shows how each controller controls the
occupant ET in a cool down scenario ($\cooldowncabin$ cabin air,
$\cooldownblock$ block temperature and $\cooldownambient$ outside
temperature). Some oscillation in ET is caused by turning on and off
the fan, due to ET's definition, which depends on air flow rate. The
RL controller cools slightly more quickly and avoids the fluctuation
in ET present in both other approaches and thus performs better overall.

\subsubsection{Effect of parameter choices\label{sub:Effect-of-parameter}}

\begin{figure}
\centering{}\includegraphics[scale=0.8]{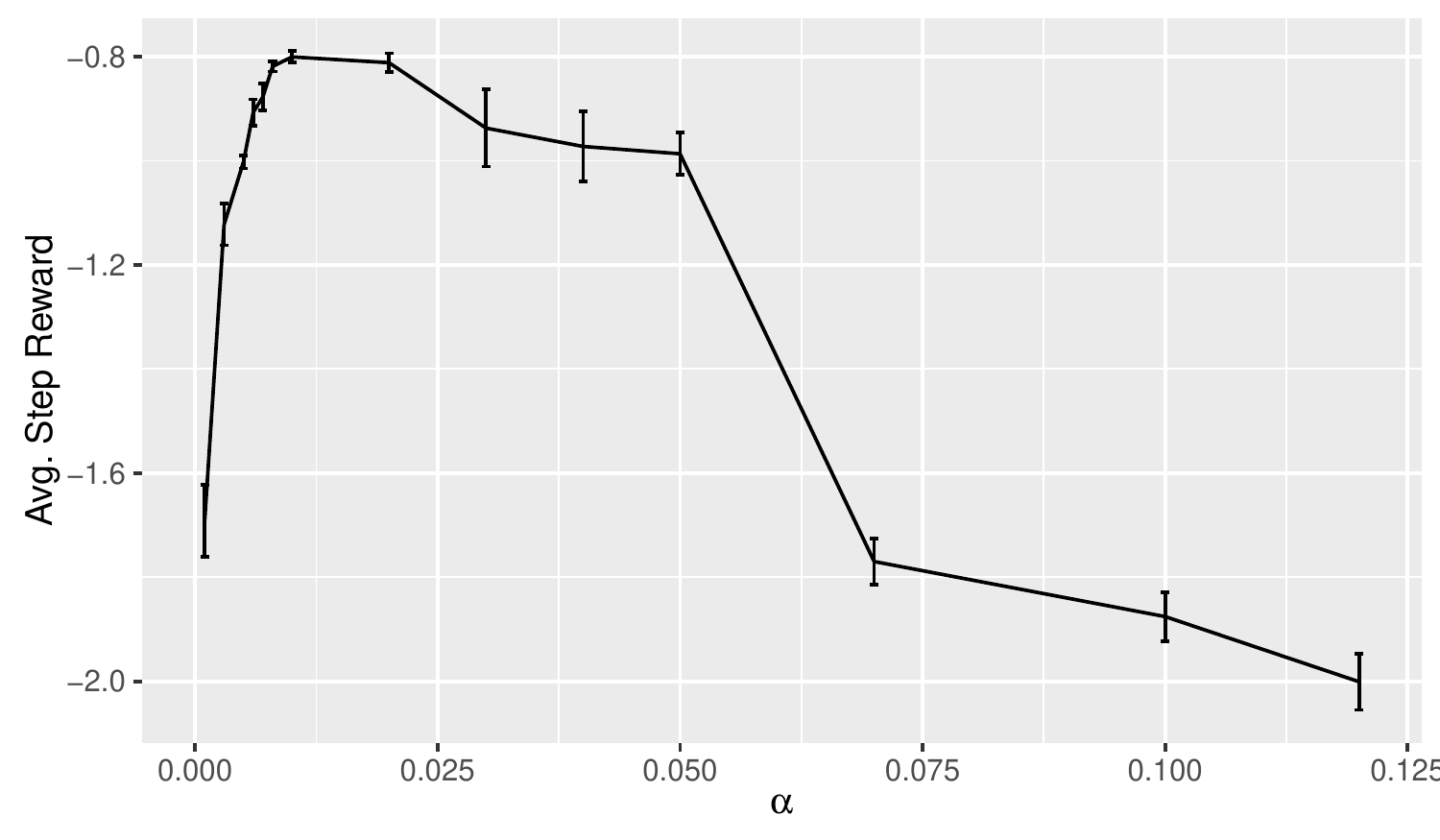}\protect\caption{Mean reward for the test scenario set obtained over episodes 1000
to 2000 shows that a small, but not too small, learning rate $\alpha=\rlbestalpha$
provides peak performance. Error bars show the two-tail 95\% confidence
interval.\label{fig:Alpha}}
\end{figure}

Learning parameters (such as $\alpha,\varepsilon,\lambda$) affect
the RL learning rate. For example, Figure~\ref{fig:Alpha} shows
how the mean reward over episodes $1\,000$--$2\,000$ changes with
the learning rate $\alpha$ and that a rate of $\rlbestalpha$ produces
the fastest learning. Similar experiments reveal best values for $\lambda$
($\rlbestlambda$) and $\varepsilon$ ($\rlbestepsilon$). Although
these parameter choices are suitable during early stages of learning,
different values may be better later on. In particular, reward performance
improves substantially if exploration is turned off $\varepsilon=0$
in the later stages of learning.

\begin{figure}

\includegraphics[scale=0.8]{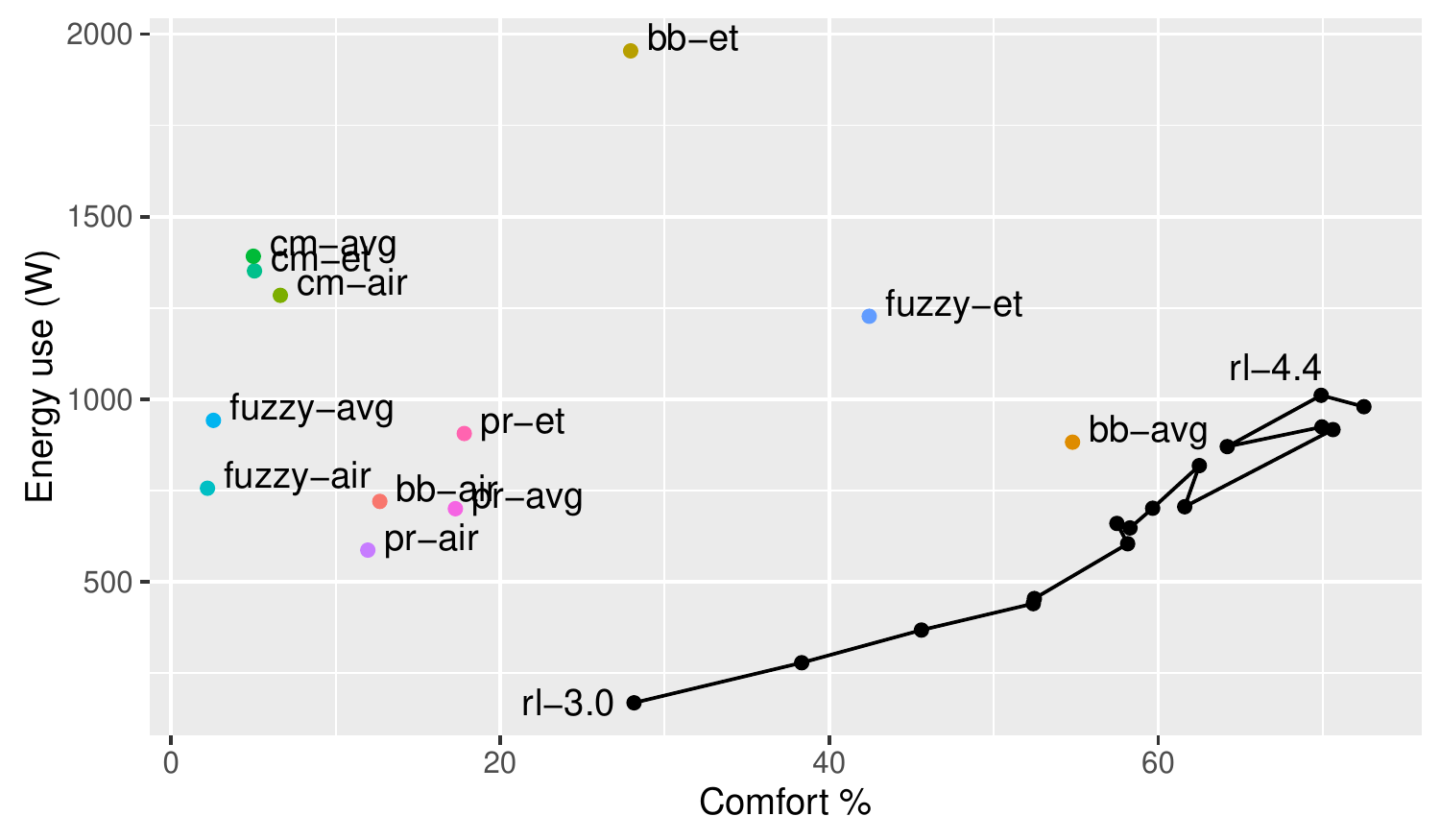}

\protect\caption{Effect of energy weight on comfort and energy showing the different
trade-offs made to either increase comfort or reduce energy use. The
black line connects results for the RL agent with an exponentially
increasing energy divisor ($\log_{10}w=[3,4.5]$ in $0.1$ steps)
and corresponds to the Pareto optimal front. \label{fig:Effect-of-energy}
Other agents are: bang-bang (bb), commercial (c), fuzzy, and proportional
(pr). The sensor type for each agent is either cabin air (air), average
of cabin and interior mass (avg), or equivalent temperature (et). }

\end{figure}

The weighting of energy versus comfort in the reward function can
make a significant difference to the performance of the resulting
policy. The tradeoff being made is reflected in Figure~\ref{fig:Effect-of-energy},
which shows the performance for policies learnt with different energy
divisor values in terms of energy use and percentage comfort. The
black line drawn in the graph corresponds to the trade-off curve (or
Pareto optimal front) and shows a progressive change in balance between
comfort and energy as the energy divisor $w$ is increased. Mostly,
comfort and energy use increases as the energy divisor $w$ is increased.
However, there is some backtracking (e.g., at $w=10^{4.3}$) that
suggests that the policy learnt for some divisors is sub-optimal.

\section{Discussion\label{sec:Discussion}}

There are several limitations of the RL controller as currently described.
First, not all factors relevant to thermal comfort are simulated or
included in the ET comfort metric, such as humidity, clothing level,
or the metabolic work rate of the subject. Of these, possibly the
most significant is humidity. Incorporating humidity into the model
could be valuable since it also helps identify window fogging and
thus allows a penalty for fogging to be included in the reward function.
If a certain thermal comfort model leads to sub-optimal comfort when
implemented as a controller, then this implies that there might be
a problem with the comfort model and thus help identify which parameter
or feature is missing. Given the diversity of opinions about comfort
models and relative importance of parameters in the literature, this
iterative approach seems best.

Second, the hand-coded controllers shown in this paper may not perform
as well as current state-of-the-art HVAC controllers. Although we
tried a commercial controller, this performs poorly in simulation.
Although this may suggest that the simulation is imperfect or that
the reward function does not take into account important factors,
it also seems likely that there is room for improvement. To understand
how much of an improvement can be obtained, side-by-side in-car comparison
is needed. 

Third, some users may prefer less fan noise, even at the expense of
being thermally uncomfortable. Furthermore, adjusting the fan speed
or recirculation setting constantly may be distracting. On the other
hand, some users prefer to hear the fan as it reassures them that
the HVAC is actively attempting to restore comfort. An advantage of
our approach is that a range of user types can be catered for by using
different reward functions with added penalties for such things as
fan noise. Note that the fact that the commercial controller performs
poorly in terms of thermal comfort could be due to a deliberate design
decision to constrain fan noise.

\subsection{Pathway to implementation in the car}

Occupant ET, which is used here as a proxy for comfort, cannot be
directly measured in a real car cabin and the need for a proxy inspired
the development of a Virtual Thermal Comfort Sensor (VTCS)~\citep{Hintea2013}.
VTCS makes use of a distributed set of sensors to estimate ET based
on a machine-learning approach. Note that all learning occurs off-line
and thus little computation is required to implement the VTCS approach
in the car. In principle, VTCS can be used to estimate comfort for
different zones such as upper and lower body as well as different
passenger positions.

A key consideration in the development of a controller is the accuracy
of the sensor. No matter how good the controller, inaccurate measurements
will lead to incorrect control. The VTCS approach has an additional
advantage that it becomes possible to integrate a set of inexpensive
sensors to accurately estimate ET rather than rely on a single sensor. 

The RL agent developed in this paper is designed to sit on top of
existing low-level controllers (such as those that control the speed
of the compressor motor). This approach has the advantage that it
makes the RL controller generic and retains any existing low-level
safety mechanisms. 

Implementing in the car provides an opportunity to receive feedback
from the end-user. This feedback might come in the form of manual
temperature adjustments. Such feedback can be incorporated as a penalty
in the reward function and thus enable some learning of preferences.
It is unclear whether learning of preferences in this way would occur
quickly enough.

\section{Conclusions and future work}

Our results show that the RL-based controller delivers better comfort
($\rlcomfort$\% time in comfort versus $\bangbangavgcomfort$\% for
the bang-bang controller with averaged sensor) more efficiently ($\rlenergy$~kW
for RL versus $\bangbangavgenergy$~kW for the bang-bang controller).
Note that the exact level of energy use may vary from this in practise
since the energy use aspect of the simulation has not been fully validated.
The performance of the RL controller is striking for two reasons:
First, the reward function does not `coach' the agent towards the
solution; reward is only provided when comfort is reached. Second,
the RL controller is not explicitly informed of the current ET but
still manages to control it in a stable way. 

There are a number of opportunities for future work. As discussed
in Section~\ref{sec:Discussion}, some of the limitations of the
approach are due to the simulator and the controller might be improved
by enhancing its realism. However, work to date on integrating with
a Dymola-based cabin simulation~\citep{Gravelle2014} has shown that
ensuring that the simulation is sufficiently fast remains a key challenge.
There are several options to improve the simulation to make it more
realistic. For example, humidity is a key factor in thermal comfort
and enables identifying screen fogging. Furthermore, a zoned approach
to the simulation would allow differential control of comfort for
different parts of the body and for different seat positions. Testing
in the car is another avenue for future work that would allow better
comparison against existing controllers. 

Actuation has become more complex with the introduction of heated
and cooled surfaces. Although it makes sense for radiant and blown-air
systems to work in concert, no current system attempts this. Similarly,
natural ventilation can be used to reduce cabin temperatures in hot
climates with minimal energy consumption. This work opens the door
to development of a holistic controller that integrates such disparate
actuators.

From the cabin HVAC designer's perspective, the RL approach raises
the abstraction level from coding boolean or fuzzy rule sets towards
making decisions about how to best model occupant comfort and its
relative importance versus noise level, screen clarity, and energy
efficiency. As this work shows, the resulting controller can be expected
to substantially improve over manually coded designs.

\section*{Acknowledgements}

The Low Carbon Vehicle Technology Project (LCVTP) was a collaborative
research project between leading automotive companies and research
partners, revolutionising the way vehicles are powered and manufactured.
The project partners included Jaguar Land Rover, Tata Motors European
Technical Centre, Ricardo, MIRA LTD., Zytek, WMG and Coventry University.
The project included 15 automotive technology development work-streams
that will deliver technological and socio-economic outputs that will
benefit the West Midlands Region. The £19 million project was funded
by Advantage West Midlands (AWM) and the European Regional Development
Fund (ERDF).

\section*{References}

\bibliographystyle{elsarticle-harv}
\addcontentsline{toc}{section}{\refname}\bibliography{AllLiterature,rl-hvac}

\end{document}